\documentclass[11pt]{article}
\usepackage{acl2016}
\usepackage{array}
\usepackage{times}
\usepackage{url}
\usepackage{latexsym}
\usepackage[usenames,dvipsnames]{color}
\usepackage{amsmath}
\usepackage{amsfonts} 
\usepackage{enumitem}
\usepackage{caption}
\usepackage{subcaption}
\usepackage{graphicx}
\usepackage[utf8]{inputenc}
\aclfinalcopy 

\newcommand*{\horzbar}{\rule[.5ex]{2.5ex}{0.5pt}}

\title{Modelling Context with User Embeddings \\ for Sarcasm Detection in Social Media}

\author{Silvio Amir \hspace{3mm}Byron C. Wallace$^{\dagger}$ \hspace{3mm}Hao Lyu$^{\dagger}$ \hspace{3mm}Paula Carvalho \hspace{3mm} Mário J. Silva\\
	   { \normalsize INESC-ID Lisboa, Instituto Superior Técnico, Universidade de Lisboa}\\
	   { \normalsize $^{\dagger}$iSchool, University of Texas at Austin}\\
	   { \small {\tt samir@inesc-id.pt~~byron.wallace@utexas.edu~~xalh8083@gmail.com }}\\
	   { \small {\tt pcc@inesc-id.pt~~mjs@inesc-id.pt }}}

\date{}

\begin{document}
\maketitle
\begin{abstract}


We introduce a deep neural network for automated sarcasm detection. Recent work has emphasized the need for models to capitalize on \emph{contextual} features, beyond lexical and syntactic cues present in utterances. For example, different speakers will tend to employ sarcasm regarding different subjects and, thus, sarcasm detection models ought to encode such speaker information. Current methods have achieved this by way of laborious feature engineering. By contrast, we propose to automatically learn and then exploit \emph{user embeddings}, to be used in concert with lexical signals to recognize sarcasm. Our approach does not require elaborate feature engineering (and concomitant data scraping); fitting user embeddings requires only the text from their previous posts. The experimental results show that our model outperforms a state-of-the-art approach leveraging an extensive set of carefully crafted features.

\end{abstract}

\section{Introduction}
\label{section:intro}

Existing social media analysis systems are hampered by their inability to accurately detect and interpret figurative language. This is particularly relevant in domains like the social sciences and politics, in which the use of figurative communication devices such as verbal irony (roughly, sarcasm) is common. Sarcasm is often used by individuals to express opinions on complex matters and regarding specific targets~\cite{carvalho2009clues}. 

\begin{figure}[t]
\centering
\includegraphics[width=1\linewidth]{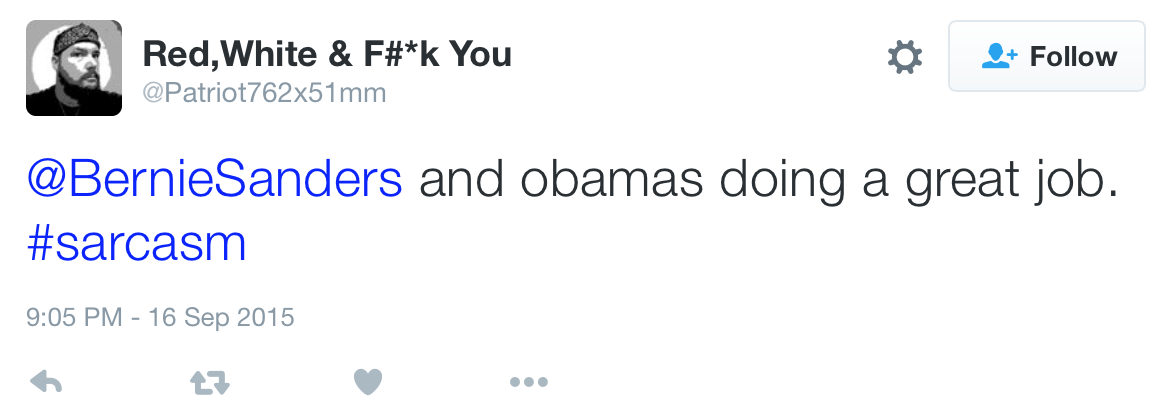}
\vspace{-1.5em}
\caption{An illustrative tweet. }
\label{fig:sarcasm_tweet}
\vspace{-1.5em}
\end{figure}

Early computational models for verbal irony and sarcasm detection tended to rely on shallow methods exploiting conditional token count regularities. But lexical clues alone are insufficient to discern ironic intent. Appreciating the \textit{context} of utterances is critical for this; even for humans~\cite{wallace-EtAl:2014:P14-2}. Indeed, the exact same sentence can be interpreted as literal or sarcastic, \emph{depending on the speaker}. Consider the sarcastic tweet in Figure \ref{fig:sarcasm_tweet} (ignoring for the moment the attached \texttt{\#sarcasm} hashtag). Without knowing the author's political leanings, it would be difficult to conclude with certainty whether the remark was intended sarcastically or in earnest.

Recent work in sarcasm detection on social media has tried to incorporate contextual information by exploiting the preceding messages of a user, to e.g., detect contrasts in sentiments expressed towards named entities \cite{khattri2015your}, infer behavioural traits \cite{rajadesingan2015sarcasm} and capture the relationship between authors and the audience \cite{bamman2015contextualized}. However, all of these approaches require the design and implementation of complex features that explicitly encode the content and (relevant) context of messages to be classified. This feature engineering is labor intensive, and depends on external tools and resources. Therefore, deploying such systems in practice is expensive, time-consuming and unwieldy. 

We propose a novel approach to sarcasm detection on social media that does not require extensive manual feature engineering. Instead, we develop a neural model that learns to represent and exploit embeddings of both \emph{content} and \emph{context}. For the former, we induce vector lexical representations via a convolutional layer; for the latter, our model learns \emph{user embeddings}. Inference concerning whether an utterance (tweet) was intended ironically (or not) is then modelled as a joint function of lexical representations and corresponding author embeddings. 

The main contributions of this paper are as follows. (i) We propose a novel convolutional neural network based model that explicitly learns and exploits user embeddings in conjunction with features derived from utterances.  (ii) We show that this model outperforms the strong baseline recently proposed by \newcite{bamman2015contextualized} by more than $2\%$ in absolute accuracy, while obviating the need to manually engineer features. (iii) We show that the learned user embeddings can capture relevant user attributes.

\section{Related Work}
\label{section:related-work} 

Verbal irony is a rhetorical device in which speakers say something other than, and often opposite to, what they actually mean.\footnote{Like other forms of subjective expression, irony and sarcasm are difficult to define precisely.} Sarcasm may be viewed as a special case of irony, where the positive literal meaning is perceived as an indirect insult \cite{dews1995not}.
 
Most of the previously proposed computational models to detect irony and sarcasm have used features similar to those used in sentiment analysis. \newcite{carvalho2009clues} analyzed comments posted by users on a Portuguese online newspaper and found that oral and gestural cues indicate irony. These included: emoticons, onomatopoeic expressions for laughter, heavy punctuation, quotation marks and positive interjections. Others have used text classifiers with features based on word and character $n$-grams, sentiment lexicons, surface patterns and textual markers~\cite{davidov2010semi,gonzalez2011identifying,reyes2013multidimensional,lukin2013really}. 
Elsewhere, \newcite{barbieri2014modelling} derived new word-frequency based features to detect irony, e.g., combinations of frequent and rare words, ambiguous words, `spoken style' words combined with `written style' words and intensity of adjectives. \newcite{riloff2013sarcasm} demonstrated that one may exploit the apparent expression of contrasting sentiment in the same utterance as a marker of verbal irony. 

The aforementioned approaches rely predominantly on features \emph{intrinsic} to texts, but these will often be insufficient to infer figurative meaning: context is needed. There have been some recent attempts to exploit contextual information, e.g. \newcite{khattri2015your} extended the notion of \textit{contrasting sentiments} beyond the textual content at hand. In particular, they analyzed previous posts to estimate the author's prior sentiment towards specific \textit{targets} (i.e., entities). A tweet is then predicted to be sarcastic if it expresses a sentiment about an entity that contradicts the author's (estimated) prior sentiment regarding the same.

\newcite{rajadesingan2015sarcasm} built a system based on theories of sarcasm expression from psychology and behavioral sciences. To operationalize such theories, they used several linguistic tools and resources (e.g. lexicons, sentiment classifiers and a PoS tagger), in addition to user profile information and previous posts, to model a range of behavioural aspects (e.g., mood, writing style). \newcite{wallace-choe-charniak:2015:ACL-IJCNLP} developed an approach for classifying posts on \emph{reddit}\footnote{\url{http://reddit.com} is a social news aggregation site comprising specific topical  \emph{sub-reddits}.} as sarcastic or literal, based in part on the interaction between the specific sub-reddit to which a post was made, the entities mentioned, and the (apparent) sentiment expressed. For example, if a post in the (politically) conservative sub-reddit mentions Obama, it is more likely to have been intended ironically than posts mentioning Obama in the progressive sub-reddit. But this approach is limited because it relies on the unique sub-reddit structure.  \newcite{bamman2015contextualized} proposed an approach that relied on an extensive, rich set of features capturing various contextual information about authors of tweets and the audience (in addition to lexical cues).  We review these at length in Section \ref{section:baselines}. 

A major downside of these and related approaches, however, is the amount of manual effort required to derive these feature sets. A primary goal of this work is to explore whether neural models can effectively learn these rich contextualizing features, thus obviating the need to manually craft them. In particular, the model we propose similarly aims to combine lexical clues with extra-linguistic information. Unlike prior work, however, our model attempts to automatically induce representations for the content and the author of a message that are predictive of sarcasm.




%


\section{Learning User Embeddings}
\label{sec:user_embeddings}

Our goal is to learn representations (vectors) that encode latent aspects of users and capture homophily, by projecting similar users into nearby regions of the embedding space. We hypothesize that such representations will naturally capture some of the signals that have been described in the literature as important indicators of sarcasm, such as contrasts between what someone believes and what they have ostensibly expressed~\cite{campbell2012there} or \newcite{kreuz1996use} principle of \textit{inferability}, stating that sarcasm requires a common ground between parties to be understood.


To induce the user embeddings, we adopt an approach similar to that described in the preliminary work of \newcite{li2015learning}. In particular, we capture relations between users and the content they produce by optimizing the conditional probability of texts, given their authors (or, more precisely, given the vector representations of their authors). This method is akin to \newcite{le2014distributed}'s \emph{Paragraph Vector} model, which jointly estimates embeddings for words and paragraphs by learning to predict the occurrence of a word $w$ within a paragraph $p$ conditioned on the (learned) representation for $p$. 



Given a sentence $S=\{w_1, \ldots, w_N\}$ where $w_i$ denotes a word drawn from a vocabulary $\mathcal{V}$, we aim to maximize the following probability:

\begin{equation}
\begin{split}
     P(S|\mathrm{user}_j) & = \sum_{w_i \in S} \log 
     P(w_i|\mathbf{u}_j) \\
                          & + \sum_{w_i \in S} \sum_{w_k \in C(w_i)} \log 
                          P(w_i|\mathbf{e}_k)  \\
\end{split}
\label{eq:user_embeddings}
\end{equation}
 
\noindent Where $C(w_i)$ denotes the set of words in a pre-specified window around word $w_i$, $\mathbf{e}_k \in \mathbb{R}^d$ and $\mathbf{u}_{j} \in \mathbb{R}^d$ denote the embeddings of word $k$ and user $j$, respectively. This objective function encodes the notion that the occurrence of a word $w$, depends both on the author of $S$ and it's neighbouring words.

The conditional probabilities in Equation \ref{eq:user_embeddings} can be estimated with log-linear models of the form:
\begin{equation}
\begin{split}
        P(w_i|\mathbf{x}) &=\frac{\exp(\mathbf{W}_i \cdot \mathbf{x} + \mathbf{b}_i)}{\sum_{k=1}^Y \exp(\mathbf{W}_k \cdot \mathbf{x}+ \mathbf{b}_k) }
\end{split}
\end{equation}
\noindent Where $\mathbf{x}$ denotes a feature vector, $\mathbf{W}_k$ and $\mathbf{b}_k$ are the weight vectors and bias for class $k$. In our case, we treat words as classes to be predicted. Calculating the denominator thus requires summing over all of the words in the (large) vocabulary, an expensive operation. To avoid this computational bottleneck, we approximate the term $P(w_i|\mathbf{e}_k)$ with~\newcite{morin2005hierarchical} Hierarchical Softmax.\footnote{As implemented in \texttt{Gensim} \cite{rehurek_lrec}.} 


To learn meaningful user embeddings, we seek representations that are predictive of individual word-usage patterns. In light of this motivation, we approximate  $P(w_i|\mathbf{u}_j)$ via the following hinge-loss objective which we aim to minimize: 

\begin{flalign}
    \nonumber \mathcal{L}(w_i,\mathrm{user}_j) &= \\ & \hspace{-3em} \sum_{w_l \in V, w_l \not{\in} S}\max(0,1 - \mathbf{e}_i \cdot \mathbf{u_j} + \mathbf{e}_l \cdot \mathbf{u_j}) 
\end{flalign}

\noindent Here, each $w_l$ (and corresponding embedding, $\mathbf{e}_l$) is a \emph{negative example}, i.e., a word not in the sentence under consideration, which was authored by user $j$. The intuition is that in the aggregate, such words are less likely to be employed by user $j$ than words observed in sentences she has authored. Thus minimizing this objective attempts to induce a representation that is discriminative with respect to word usage. 

In practice, $V$ will be very large and hence we approximate the objective via \emph{negative sampling}, a variant of Noise Contrastive Estimation. The idea is to approximate the objective function in a binary classification task by learning to discriminate between observed positive examples (sampled from the true distribution) and \emph{pseudo}-negative examples (sampled from a large space of predominantly negative instances). Intuitively, this shifts probability mass to plausible observations. See \newcite{dyer2014notes} for notes on Negative Sampling and Noise Contrastive Estimation.

Previous work by \newcite{collobert2011natural} showed that this approach works well in representation learning tasks when a sufficient amount of training data is available . However, we have access to only a limited amount of text for each user (see Section \ref{sec:experiments}). We hypothesize that this problem can be alleviated by carefully selecting the negative samples that mostly contribute to `push' the vectors into the appropriate region of the embedding space (i.e., closer to the words commonly employed by a given user and far from other words). This requires designing a strategy for selectively sampling negative examples. One straightforward approach would be to sample from a user-specific unigram model, informing which words are less likely to be utilized by that user. But estimating the parameters of such model with scarce data would be prone to overfitting. Instead, we sample from a unigram distribution estimated from the posts authored by all the users. The goal is to select the most commonly used words as the negative samples, thereby forcing the representations to capture the differences between the words a given individual employs and the words that everyone commonly uses.

\section{Proposed Model}
\label{sec:model}

\begin{figure*}[htb]
\centering
\includegraphics[width=0.75\linewidth]{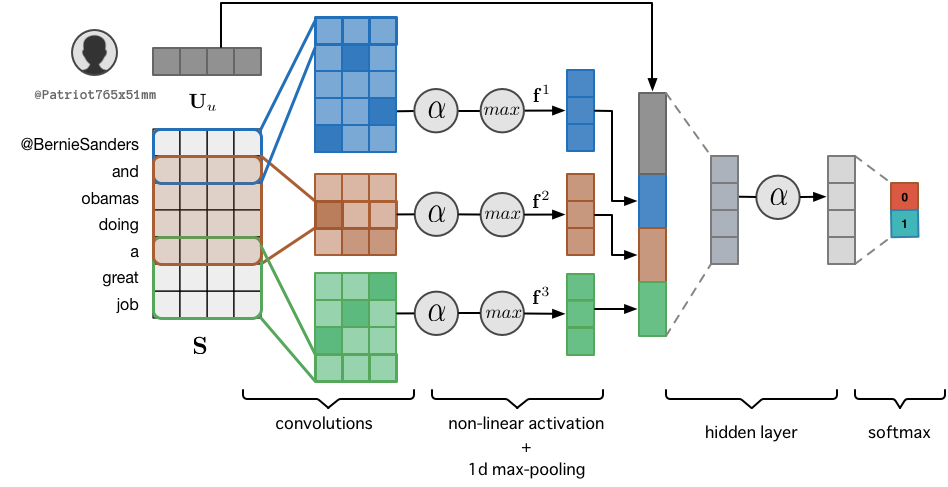}
\caption{\label{fig:model} Illustration of the CUE-CNN model for sarcasm detection. The model learns to represent and exploit embeddings of both \emph{content} and \emph{users} in social media.}
\end{figure*}

We now present the details of our proposed sarcasm detection model. Given a message $S$ authored by user $u$, we wish to capture both the relevant aspects of the \textit{content} and the relevant \textit{contextual} information about the author. To represent the content, we use pre-trained word embeddings as the input to a convolutional layer that extracts high-level features. More formally, let $\mathbf{E} \in \mathbb{R}^{d \times |\mathcal{V}|}$ be a pre-trained word embedding matrix, where each column represents a word from the vocabulary $\mathcal{V}$ as a $d$ dimensional vector. By selecting the columns of $\mathbf{E}$ corresponding to the words in $S$, we form the sentence matrix:

\begin{equation}
\mathbf{S} = \left[
  \begin{array}{ccc}
    \horzbar & \mathbf{e}_1 & \horzbar \\
             & \vdots       &          \\
    \horzbar & \mathbf{e}_m & \horzbar
  \end{array}
\right]
\label{eq:sentence}
\end{equation}

A convolutional layer is composed of a set of filters $\mathbf{F} \in \mathbb{R}^{d\times h}$ where $h$ is the \textit{height} of the filter. Filters \textit{slide} across the input, extracting $h$-gram features that constitute a feature map $\mathbf{m} \in \mathbb{R}^{|S|-h+1}$, where each entry is obtained as
\begin{equation}
\mathbf{m}_i = \alpha(\mathbf{F}\cdot \mathbf{S}_{[i:i-h+1]} + b)
\end{equation}
\noindent with $i=1,\ldots i-h+1$. Here, $\mathbf{S}_{[i:j]}$ denotes a sub-matrix of $\mathbf{S}$ (from row $i$ to row $j$), $b \in \mathbb{R}$ is an additive bias and $\alpha(\cdot)$ denotes a non-linear activation function, applied element-wise. We transform the resultant feature map into a scalar using \emph{max-pooling}, i.e., we extract the largest value in the map.
We use 3 filters (with varying heights) each of which generates $M$ feature maps that are reduced to a vector ${\mathbf{f}^k = [max(\mathbf{m}^1)~\oplus max(\mathbf{m}^2) \ldots \oplus~max(\mathbf{m}^{M})]}$, where $\oplus$ denotes concatenation. We set $\alpha(\cdot)$ to be the \textit{Rectified Linear Unit} activation function~\cite{nair2010rectified}. The output of all the filters is then combined to form the final representation $\mathbf{c}=[\mathbf{f}^1 \oplus \mathbf{f}^2 \oplus \mathbf{f}^3]$. We will denote this feature vector of a specific sentence $S$ by $\mathbf{c}_S$. 

To represent the context, we assume there is a user embedding matrix $\mathbf{U} \in \mathbb{R}^{d \times N}$, where each column represents one of $N$ users with a $d$ dimensional vector. The parameters of this embedding matrix can be initialized randomly or using the approach described in Section \ref{sec:user_embeddings}. Then, we simply map the author of the message into the user embedding space by selecting the corresponding column of $\mathbf{U}$. Letting $\mathbf{U}_u$ denote the user embedding of author $u$, we formulate our sarcasm detection model as follows:

\begin{equation}
\begin{split}
    P(y=k|s,u;\mathbf{\theta}) & \propto \mathbf{Y}_k \cdot g(\mathbf{c}_S \oplus \mathbf{U}_u) + \mathbf{b}_k \\
    g(\mathbf{x}) &= \alpha(\mathbf{H} \cdot  \mathbf{x} + \mathbf{h})
\end{split}
\end{equation}
\noindent where $g(\cdot)$ denote the activations of a hidden layer, capturing the relations between the content and context representations, and ${\theta=\{\mathbf{Y},\mathbf{b},\mathbf{H},\mathbf{h},\mathbf{F}^1,\mathbf{F}^2,\mathbf{F}^3,\mathbf{E},\mathbf{U}\}}$ are parameters to be estimated during training. Here, $\mathbf{Y} \in \mathbb{R}^{2\times z}$ and $\mathbf{b} \in \mathbb{R}^2$ are the weights and bias of the output layer; $\mathbf{H} \in \mathbb{R}^{z \times 3M+d}$ and $\mathbf{h} \in \mathbb{R}^{z}$ are the weights and bias of the hidden layer; and $\mathbf{F}^i$ are the weights of the convolutional filters.  Henceforth, we will refer to this approach as \textit{Content and User Embedding Convolutional Neural Network} (CUE-CNN). Figure \ref{fig:model} provides an illustrative schematic depicting this model.

\section{Experimental Setup}
\label{sec:experiments}

We replicated \newcite{bamman2015contextualized} experimental setup using (a subset of) the same Twitter corpus. The labels were inferred from self-declarations of sarcasm, i.e., a tweet is considered sarcastic if it contains the hashtag \texttt{\#sarcasm} or \texttt{\#sarcastic} and deemed non-sarcastic otherwise.\footnote{Note that this is a form of noisy supervision, as of course all sarcastic tweets will not be explicitly flagged as such.} To comply with Twitter terms of service, we were given only the tweet ids along with the corresponding labels and had to retrieve the messages ourselves. By the time we tried to retrieve the messages, some of them were not available. We also did not have access to the historical user tweets used by Bamman and Smith, hence, for each author and mentioned user, we scraped additional tweets from their Twitter feed. Due to restrictions in the Twitter API, we were only able to crawl at most $1000$ \textit{historical tweets} per user.\footnote{The original study \cite{bamman2015contextualized} was done with at most $3,200$ historical tweets.}  Furthermore, we were unable to collect historical tweets for a significant proportion of the users, thus, we discarded messages for which no contextual information was available, resulting in a corpus of $11,541$ tweets involving $12,500$ unique users (authors and mentioned users). It should also be noted that our historical tweets were posted \textit{after} the ones in the corpus used for the experiments. 

\subsection{Baselines}
\label{section:baselines} 

We reimplemented \newcite{bamman2015contextualized}'s sarcasm detection model. This a simple, logistic-regression based classifier that exploits rich feature sets to achieve strong performance. These are detailed at length in the original paper, but we briefly summarize them here:

\begin{itemize}[itemsep=0pt]
    \item \textbf{tweet-features}, encoding attributes of the target tweet text, including: uni- and bi-gram bag of words (BoW) features; \newcite{brown1992class} word clusters indicators; unlabeled dependency bigrams (both BoW and with Brown cluster representations); part-of-speech, spelling and abbreviation features; inferred sentiment, at both the tweet and word level; and `intensifier' indicators. 
    
    \item \textbf{author-features}, aimed at encoding attributes of the author, including: historically `salient' terms used by the author; the inferred distribution over topics\footnote{The topics were extracted from Latent Dirichlet Allocation \cite{blei2003latent}.} historically tweeted about by the user; inferred sentiment historically expressed by the user; and author profile information (e.g., profile BoW features).
    
    \item \textbf{audience-features}, capturing properties of the \emph{addressee} of tweets, in those cases that a tweet is directed at someone (via the @ symbol). A subset of these, duplicate the aforementioned author features for the addressee. Additionally, author/audience interaction features are introduced, which capture similarity between the author and addressee, w.r.t. inferred topic distributions. Finally, this set includes a feature capturing the frequency of past communication between the author and addressee. 
    
    
    \item \textbf{response-features}, for tweets written in response to another tweet. This set of features captures information relating the two, with BoW features of the original tweet and pairwise cluster indicator features, which the encode Brown clusters observed in both the original and response tweet.
    
\end{itemize}

We emphasize that implementing this rich set of features took considerable time and effort. This motivates our approach, which aims to effectively induce and exploit contextually-aware representations without manual feature engineering.

To assess the importance of modelling contextual information in sarcasm detection, we considered two groups of models as baselines: the first only takes into account the content of a target tweet. The second, combines lexical cues with contextual information. The first group includes the following models:

\begin{itemize}[itemsep=0pt]
    \item \textsc{Unigrams}: $\ell_2$-regularized logistic regression classifier with binary unigrams as features.
    \item \textsc{Tweet Only}: $\ell_2$-regularized logistic regression classifier with binary unigrams and  \textbf{tweet-features}.
    \item \textsc{nBOW}: Logistic regression with neural word embeddings as features. Given a sentence matrix $\mathbf{S}$ (Eq.~\ref{eq:sentence}) as input, a $d$-dimensional feature vector is computed by summing the individual word embeddings.
    \item \textsc{NLSE}: The Non-linear subspace embedding model due to~\newcite{astudillo-EtAl:2015:ACL-IJCNLP}. The NLSE model adapts pre-trained word embeddings for specific tasks by learning a projection into a small subspace. The idea is that this subspace captures the most discriminative latent aspects encoded in the word embeddings. Given a sentence matrix $\mathbf{S}$, each word vector is first projected into the subspace and then transformed through an element-wise sigmoid function. The final sentence representation is obtained by summing the (adapted) word embeddings and passed into a softmax layer that outputs the predictions.
    \item \textsc{Cnn}: The CNN model for text classification proposed by \newcite{kim:2014:EMNLP2014}, using only features extracted from the convolutional layer acting on the lexical content. The input layer was initialized with pre-trained word embeddings.
\end{itemize}

The second group of baselines consists of the following models:

\begin{itemize}[itemsep=0pt]
    \item \textsc{Tweet+*}: $\ell_2$-regularized logistic regression classifier with a combination of \textbf{tweet-features} and each of the aforementioned \newcite{bamman2015contextualized} feature sets.
    \item \textsc{Shallow CUE-CNN}: A simplified version of our neural model for sarcasm detection, without the hidden layer. We evaluated two variants: initializing the user embeddings at random, and initializing the user embeddings with the approach described in Section  \ref{sec:user_embeddings} (\textsc{Shallow CUE-CNN+user2vec}). In both cases, the (word and user) embeddings weights were updated during training.
    \item \textsc{CUE-CNN+*}: Our proposed neural network for sarcasm detection. We also evaluated the two aforementioned variants: randomly initialized user embeddings and pre-trained user embeddings. But here we compared two different approaches for the negative sampling procedure, namely, sampling from a unigram  distribution (\textsc{CUE-CNN+user2vec}) and sampling uniformly at random from the vocabulary (\textsc{CUE-CNN+user2vec-UnifRand}). 
    
\end{itemize}

\subsection{Pre-Training Word and User Embeddings}

We first trained \newcite{mikolov2013distributed}'s \emph{skip-gram} model variant to induce word embeddings using the union of: \newcite{owoputi2013improved}'s dataset of 52 Million unlabeled tweets, \newcite{bamman2015contextualized} sarcasm dataset and 5 Million historical tweets collected from users.


To induce user embeddings, we estimated a unigram distribution with maximum likelihood estimation. Then, for each word in a tweet, we extracted $15$ negative samples (for the first term in Eq.\ref{eq:user_embeddings}) and used the skip-gram model to pre-compute the conditional probabilities of words occurring in a window of size $5$ (for the second term in Eq.\ref{eq:user_embeddings}). Finally, Equation \ref{eq:user_embeddings} was minimized via Stochastic Gradient Descent  on $90\%$ of the historical data, holding out the remainder for validation and using the $P({\text{tweet text}}|{\text{user}})$ as early stopping criteria. 


Note that the parameters for each user only depend on their own tweets; this allowed us to perform these computations in parallel to speed-up the training.

\begin{figure*}[bt]
\centering
\begin{subfigure}{.49\linewidth}
  \centering
  \includegraphics[width=1\linewidth]{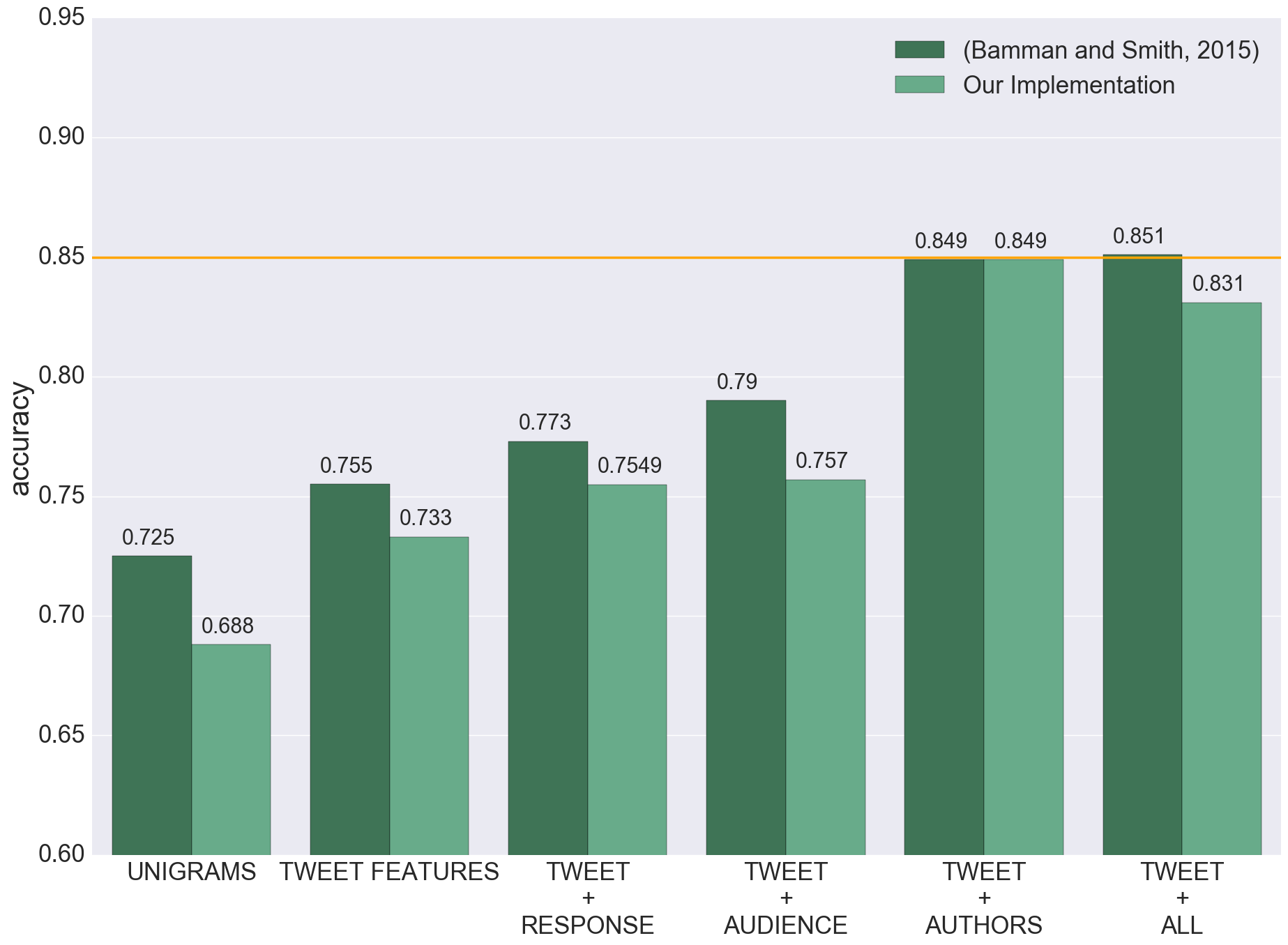}
  \subcaption{\label{fig:sub1}Performance of the linear classifier baselines. We include the results reported by \newcite{bamman2015contextualized} as a reference. Discrepancies between their reported results and those we achieved with our re-implementation reflect the fact that their experiments were performed using a significantly larger training set and more historical tweets than we had access to.~\\}
  \end{subfigure}%
\hfill
\begin{subfigure}{.49\linewidth}
  \centering
  \includegraphics[width=1\linewidth]{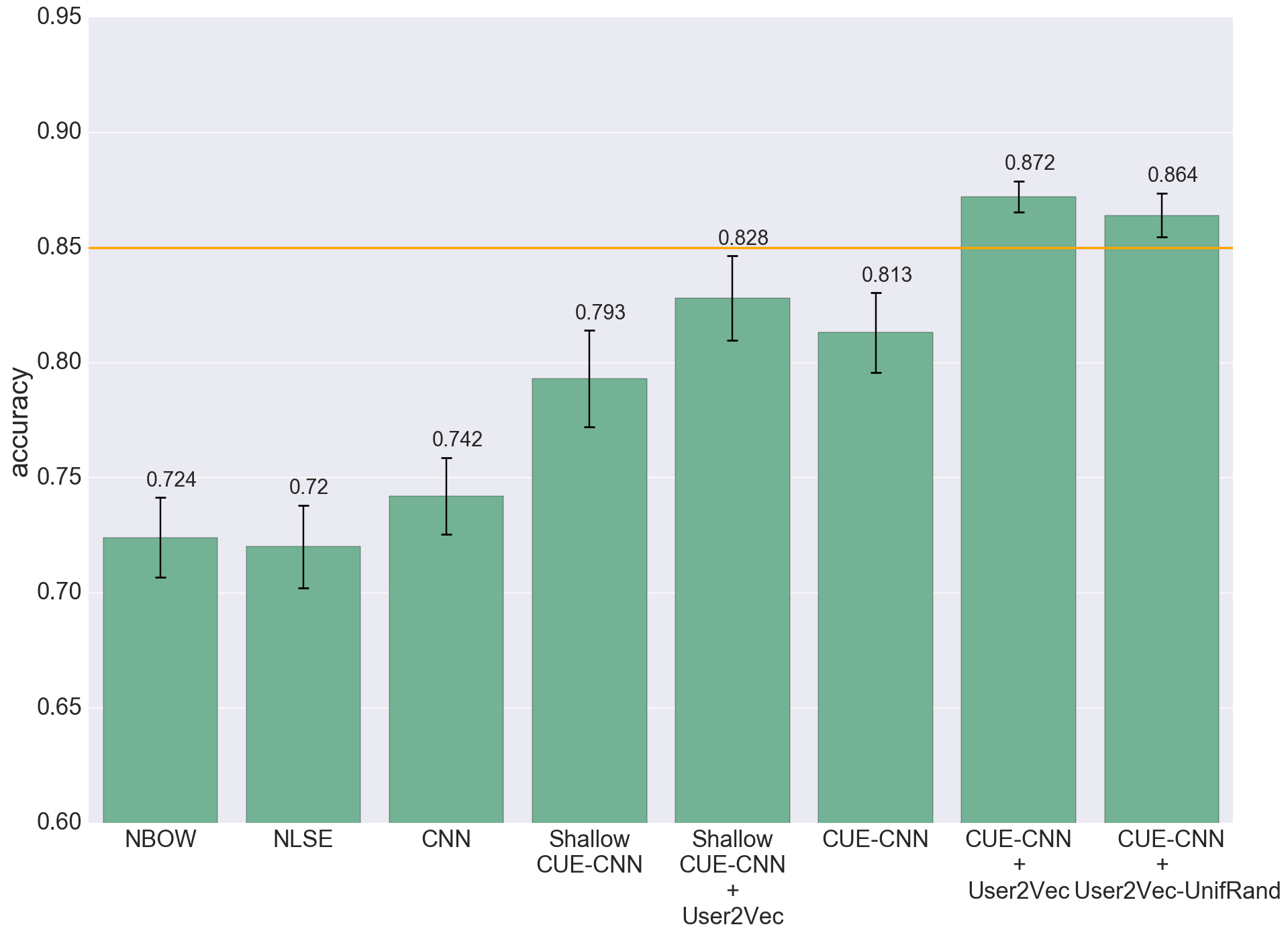}
  \subcaption{\label{fig:sub2}Performance of the proposed neural models. We compare simple neural models that only consider the lexical content of a message (first 3 bars) with architectures that explicitly model the context. Bars 4 and 5 show the gains obtained by pre-training the user embeddings. The last 2 bars compare different negative sampling procedures for the user embedding pre-training.}
\end{subfigure}
\caption{Comparison of different models. The left sub-plot shows linear model performance; the right shows the performance of neural variants. The horizontal line corresponds to the best performance achieved via linear models with rich feature sets. Performance was measured in terms of average accuracy over 10-fold cross-validation; error bars depict the variance.}
\label{fig:results}
\end{figure*}
m
\subsection{Model Training and Evaluation}

Evaluation was performed via 10-fold cross-validation. For each split, we fit models to $80\%$ of the data, tuned them on $10\%$ and tested on the remaining, held-out $10\%$. These data splits were kept fixed in all the experiments. For the linear classifiers, in each split, the regularization constant was selected with a linear search over the range $C = [1e^{-4},1e^{-3}, 1e^{-2}, 1e^{-1}, 1, 10]$ using the training set to fit the model and evaluating on the tuning set. After selecting the best regularization constant, the model was re-trained on the union of the train and tune sets, and evaluated on the test set.

To train our neural model, we first had to choose a suitable architecture and hyperparameter set. However, selecting the optimal network parametrization would require an extensive search over a large configuration space. Therefore, in these experiments, we followed the recommendations in \newcite{zhang2015sensitivity}, focusing our search over combinations of dropout rates ${D=[0.0, 0.1, 0.3, 0.5]}$, filter heights ${H=[(1,3,5),(2,4,6),(3,5,7),(4,6,8),(5,7,9)]}$, number of feature maps ${M=[100,200,400,600]}$ and size of the hidden layer ${Z=[25,50,75,100]}$. 

We performed random search by sampling without replacement over half of the possible configurations. For each data split, 20\% of the training set was reserved for early stopping. We compared the sampled configurations by fitting the model on the remaining training data and testing on the tune set. After choosing the best configuration, we re-trained the model on the union of the train and tune set (again reserving 20\% of the data for early stopping) and evaluated on the test set. 

The model was trained by minimizing the cross-entropy error between the predictions and true labels, the gradients w.r.t to the network parameters were computed with backpropagation ~\cite{rumelhart1988learning} and the model weights were updated with the AdaDelta rule~\cite{zeiler2012adadelta}.

\begin{figure*}[tb]
\centering
\begin{subfigure}{.49\linewidth}
  \centering
  \includegraphics[width=1\linewidth]{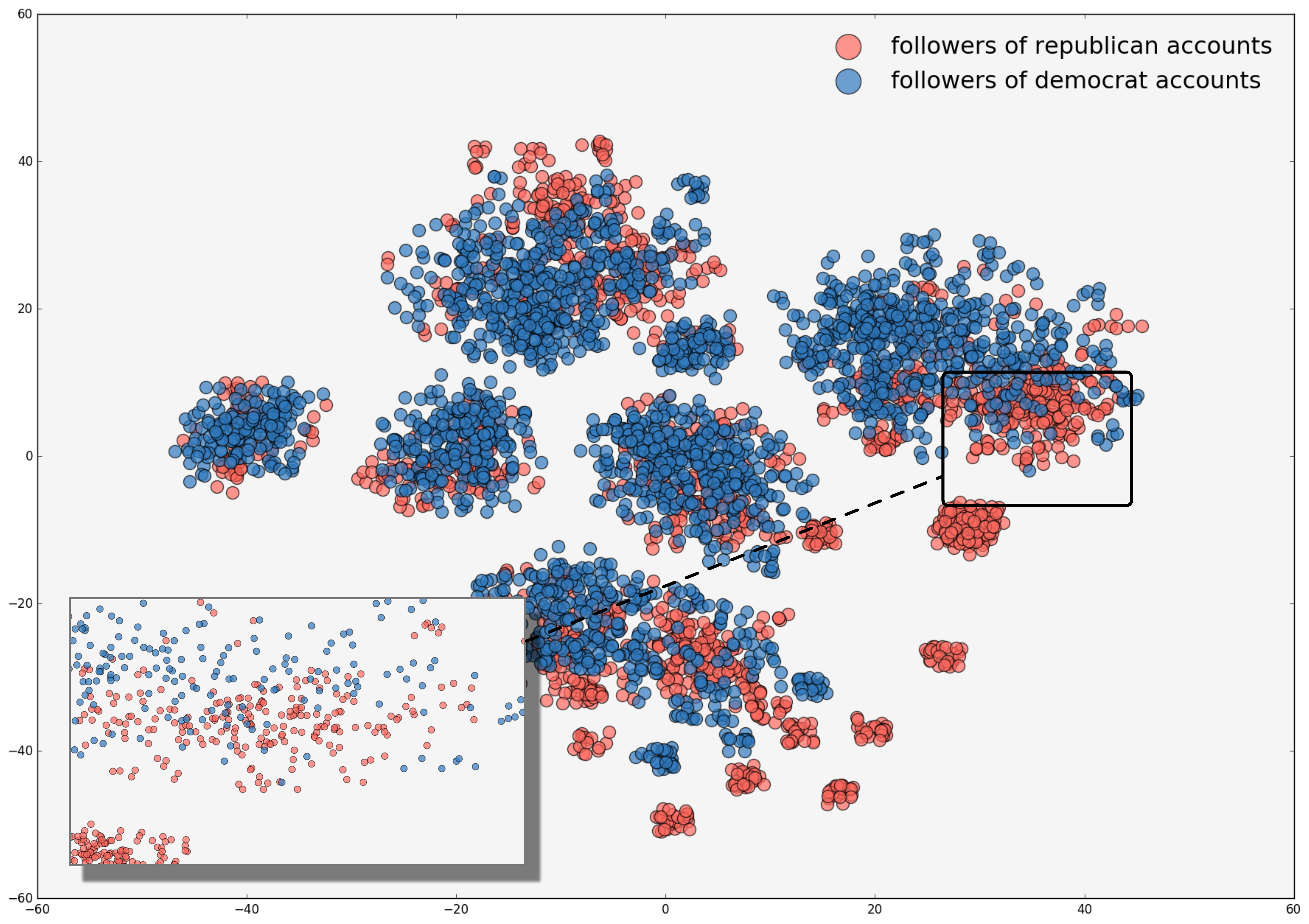}
  \subcaption{Users colored according to the politicians they follow on Twitter: the blue circles represent users that follow at least one of the (democrats) accounts: \textit{@BarackObama}, \textit{@HillaryClinton} and \textit{@BernieSanders}; the red circles represent users that follow at least one of the (republicans) accounts: \textit{@marcorubio}, \textit{@tedcruz} and \textit{@realDonaldTrump}. Users that follow accounts from both groups were excluded. We can see that users with a similar political leaning tend to have similar vectors.}
\label{fig:user_embs_politics}
\end{subfigure}%
\hfill
\begin{subfigure}{.49\linewidth}
  \centering
  \includegraphics[width=1\linewidth]{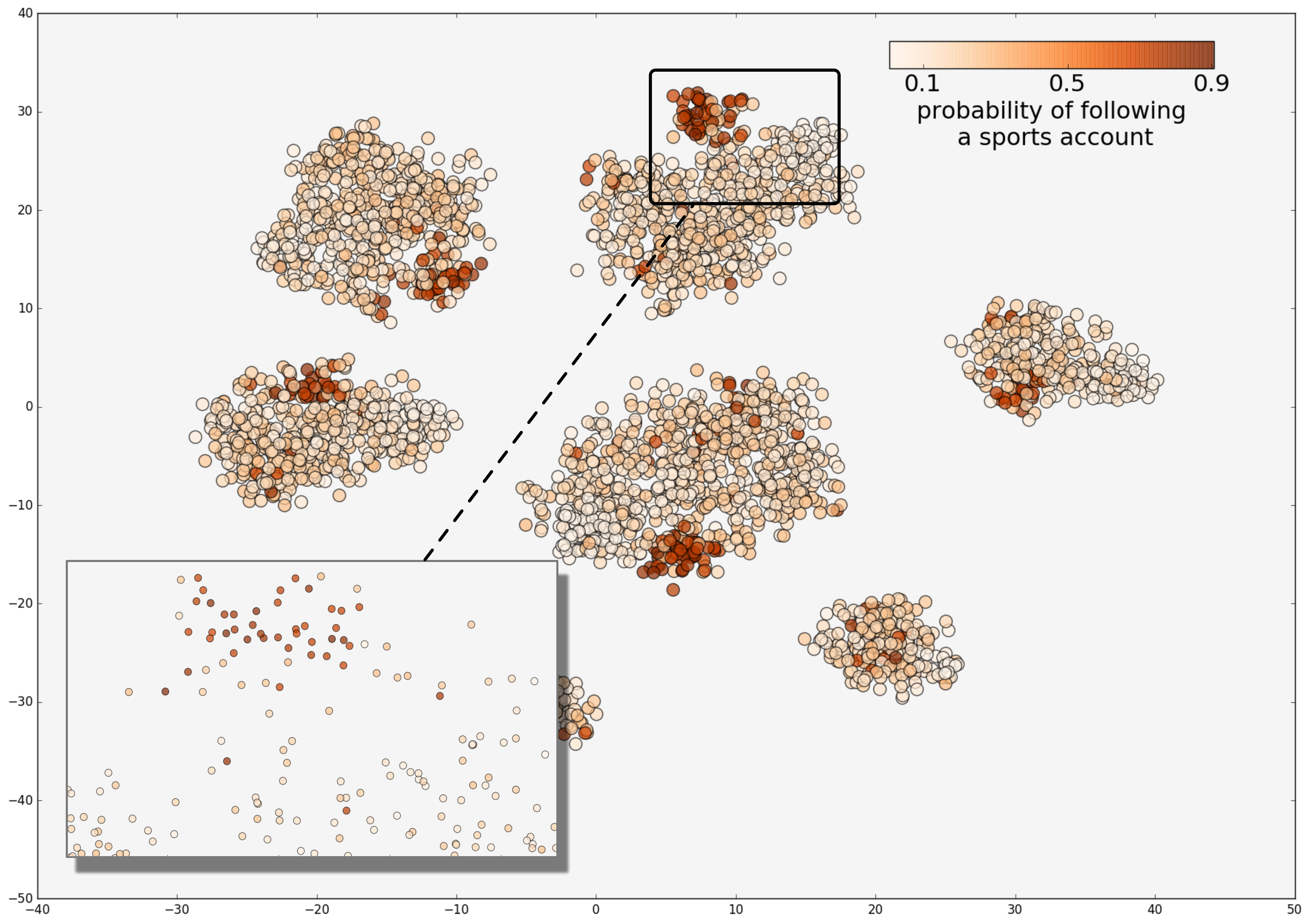}
  \subcaption{Users colored with respect to the likelihood of following a sports account. The 500 most popular accounts (according to the authors in our training data) were manually inspected and 100 sports related accounts were selected, e.g., \textit{@SkySports}, \textit{@NBA} and \textit{@cristiano}. We should note that users for which the probabilities lied in the range between $0.3-0.7$ were discarded to emphasize the extremes.~\\}
  \label{fig:user_embs_sports}
\end{subfigure}
\caption{T-SNE projection of the user embeddings into 2-dimensions. The users are color coded according to their political preferences and interest in sports. The visualization suggests that the learned embeddings capture some notion of homophily.}
\label{fig:user_embs}
\end{figure*}

\section{Results}
\label{section:results}

\subsection{Classification Results}

Figure \ref{fig:results} presents the main experimental results. In Figure \ref{fig:sub1}, we show the performance of linear classifiers with the manually engineered feature sets proposed by \newcite{bamman2015contextualized}. Our results differ slightly from those originally reported. Nonetheless, we observe the same general trends: namely, that including contextual features significantly improves the performance, and that the biggest gains are attributable to features encoding information about the authors of tweets.

The results of neural model variants are shown in Figure \ref{fig:sub2}. Once again, we find that modelling the context (i.e., the author) of a tweet yields significant gains in accuracy. The difference is that here the network jointly \emph{learns} appropriate user representations, lexical feature extractors and, finally, the classification model. Further improvements are realized by pre-training the user embeddings (we elaborate on this in the following section). We see additional gains when we introduce a hidden layer capturing the \emph{interactions} between the context (i.e., user vectors) and the content (lexical vectors). This is intuitively agreeable: the recognition of sarcasm is possible when we jointly consider the speaker and the utterance at hand. Interestingly, we observed that our proposed model not only outperforms all the other baselines, but also shows less variance over the cross-validation experiments.

Finally, we compared the effect of obtaining negative samples uniformly at random with sampling from a unigram distribution. The experimental results show that the latter improves the accuracy of the model by $0.8\%$. We believe the reason is that considering the most likely words (under the model) as negative samples, helps by pushing the user vectors away from non-informative words and simultaneously closer to the most discriminative words for that user.

\begin{figure}[bht]
\centering
\includegraphics[width=1\linewidth]{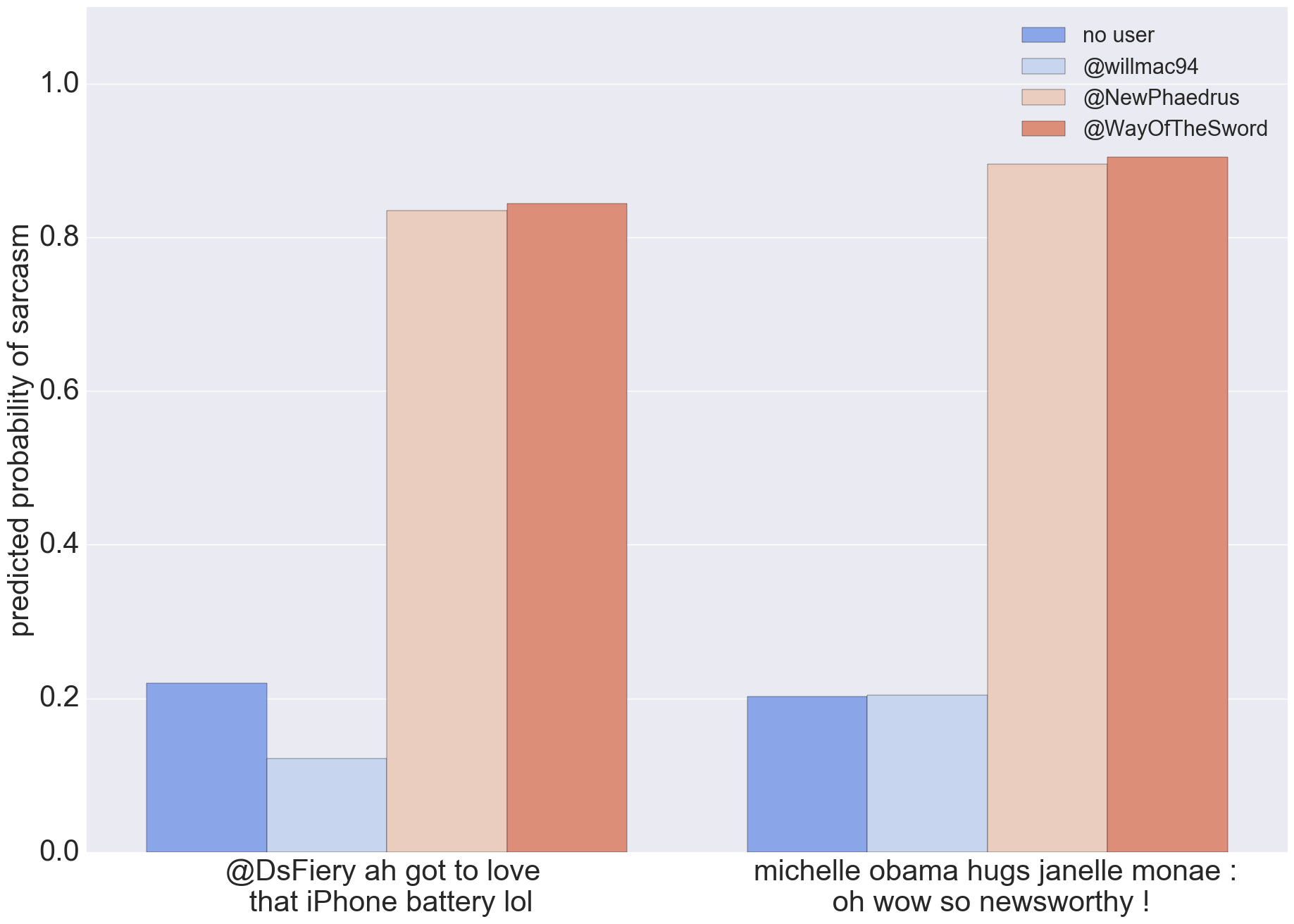}
\caption{\label{fig:probs} Two sarcastic examples that were misclassified by a simple CNN (\texttt{no user}). Using the CUE-CNN with contextual information drastically changes the model's predictions on the same examples.}
\end{figure}

\subsection{User Embedding Analysis}

We now investigate the user embeddings in more detail. In particular, we are interested in two questions: first, what aspects are being captured in these representations; and second, how they contribute to the improved performance of our model.

To investigate the first question, we plotted a T-SNE projection \cite{maaten2008visualizing} of the high-dimensional vector space where the users are represented into two-dimensions. We then colored each point (representing a user) according to their apparent political leaning (Figure \ref{fig:user_embs_politics}), and according to their interest in sports (Figure \ref{fig:user_embs_sports}). These attributes were inferred using the Twitter accounts that a user follows, as a proxy. The plots suggest that the user vectors are indeed able to capture latent aspects, such as political preferences and personal interests. Moreover, the embeddings seem to uncover a notion of homophily, i.e. similar users tend to occupy neighbouring regions of the embedding space. Regarding the second question, we examined the influence of the contextual information on the model's predictions. To this end, we measured the response of our model to the same textual content with different hypothetical contexts (authors). We selected two examples that were misclassified by a simple CNN and ran them trough the CUE-CNN model with three different user embeddings. In Figure \ref{fig:probs}, we show these examples along with the predicted probabilities of being a sarcastic post, when no user information is considered and when the author is taken into account. We can see that the predictions drastically change when contextual information is available and that two of the authors trigger similar responses on both examples. This example provides evidence that our model captures the intuition that the same utterance can be interpreted as sarcastic or not, depending on the speaker.

\section{Conclusions}

We have introduced CUE-CNN, a novel, deep neural network for automatically recognizing sarcastic utterances on social media. Our model jointly learns and exploits embeddings for the content and users, thus integrating information about the speaker and what he or she has said. This is accomplished without manual feature engineering. Nonetheless, our model \emph{outperforms} (by over 2\% in absolute accuracy) a recently proposed state-of-the-art model that exploits an extensive, hand-crafted set of features encoding user attributes and other contextual information. Unlike other approaches that explicitly exploit the structure of particular social media services, such as the forum where a message was posted or metadata about the users, learning user embeddings only requires their preceding messages. Yet, the obtained vectors are able to capture relevant user attributes and a soft notion of homophily.  This, we believe, makes our model easier to deploy over different social media environments. 

Our implementation of the proposed method and the datasets used in this paper have been made publicly available\footnote{\url{https://github.com/samiroid/CUE-CNN}}. As future work, we intended to further explore the user embeddings for context representation, namely by also incorporating the interaction between the author and the audience into the model.




\section*{Acknowledgments}

This work was supported in part by the Army Research Office (grant W911NF-14-1-0442) and by The Foundation for Science and Technology, Portugal (FCT), through contracts UID/CEC/50021/2013, EXCL/EEI-ESS/0257/2012 (DataStorm), grant UTAP-EXPL/EEIESS/0031/2014 and Ph.D. scholarship SFRH/BD/89020/2012. This work was also made possible by the support of the Texas Advanced Computer Center (TACC) at UT Austin.

\bibliography{acl2016}
\bibliographystyle{acl2016}

\end{document}